\def\CLASSINPUTinnersidemargin{0.75in}
\def\CLASSINPUToutersidemargin{0.75in}
\def\CLASSINPUTtoptextmargin{0.75in}
\def\CLASSINPUTbottomtextmargin{0.75in}

\documentclass[letterpaper, 10pt, conference]{IEEEtran}  

\IEEEoverridecommandlockouts                              

\def\IEEEtitletopspaceextra{0.25in}

\usepackage{graphicx}
\usepackage{amsmath, amssymb, amsfonts}
\usepackage{textcomp}
\usepackage{multirow}
\usepackage{cite}
\usepackage[table, xcdraw]{xcolor}
\usepackage{hyperref}
\hypersetup{hidelinks}
\usepackage{siunitx}

\usepackage{makecell}
\usepackage{romannum}
\usepackage{enumerate}
\usepackage{enumitem}
\usepackage{wrapfig}

\usepackage{algorithm}
\usepackage[ruled,vlined,algo2e]{algorithm2e}
\usepackage{algpseudocode}
\usepackage{amsmath}
\algrenewcommand\algorithmiccomment[1]{\hfill// #1}
\algrenewcommand\alglinenumber[1]{}

\usepackage{xspace}

\definecolor{deep-red}{RGB}{192, 0, 0}
\definecolor{deep-purple}{RGB}{120, 0, 170}
\definecolor{good-green}{RGB}{0,135,0} 
\definecolor{purple}{RGB}{210, 0, 210}

\title{\LARGE \bf
SurgSync: Time-Synchronized Multi-Modal Data Collection Framework and Dataset for Surgical Robotics
}

\author{$^{*}$Haoying Zhou$^{1,2}$, $^{*}$Chang Liu$^{2}$, Yimeng Wu$^{2}$, Junlin Wu$^{2,3}$,  Zijian Wu$^{4}$, Yu Chung Lee$^{4}$, Sara Martuscelli$^{5}$, \\Septimiu E. Salcudean$^{4}$, Gregory S. Fischer$^{1}$ and Peter Kazanzides$^{2,3}$

\thanks{This work was supported in part by NSF AccelNet awards OISE-1927275 and OISE-1927354.}
\thanks{$^{*}$These authors contributed equally to this work.}
\thanks{$^{1}$Department of Robotics Engineering, Worcester Polytechnic Institute, Worcester, MA, USA. Emails: \texttt{{hzhou6, gfischer}@wpi.edu}}
\thanks{$^{2}$Laboratory for Computational Sensing and Robotics, Johns Hopkins University, Baltimore, MD, USA.}
\thanks{$^{3}$Department of Computer Science, Johns Hopkins University, Baltimore, MD, USA. Email: \texttt{pkaz@jhu.edu}}
\thanks{$^{4}$ Robotics and Control Laboratory, the University of British Columbia, Vancouver, Canada.}
\thanks{$^{5}$ Department of Electronics, Information and Bioengineering, Politecnico di Milano, Milano, Italy}
}

\def\BibTeX{{\rm B\kern-.05em{\sc i\kern-.025em b}\kern-.08em T\kern-.1667em\lower.7ex\hbox{E}\kern-.125emX}} 

\begin{document}

\maketitle


\begin{abstract}

Most existing robotic surgery systems adopt a human-in-the-loop paradigm, often with the surgeon directly teleoperating the robotic system. Adding intelligence to these robots would enable higher-level control, such as supervised autonomy or even full autonomy. However, artificial intelligence (AI) requires large amounts of training data, which is currently lacking. This work proposes SurgSync, a multi-modal data collection framework with offline and online synchronization to support training and real-time inference, respectively. The framework is implemented on a da Vinci Research Kit (dVRK) and introduces (1) dual-mode (online/offline-matching) synchronized recorders, (2) a modern stereo endoscope to achieve image quality on par with clinical systems, and (3) additional sensors such as a side-view camera and a novel capacitive contact sensor to provide ground truth contact data. The framework also incorporates a post-processing toolbox for tasks such as depth estimation, optical flow, and a practical kinematic reprojection method using Gaussian heatmap. User studies with participants of varying skill levels are performed with ex-vivo tissue to provide clinically realistic data, and a network for surgical skill assessment is employed to demonstrate utilization of the collected data. Through the user study experiments, we obtained a dataset of 214 validated instances across multiple canonical training tasks.  All software and data are available at \href{https://surgsync.github.io/}{surgsync.github.io}.
\end{abstract}

\section{INTRODUCTION}

\begin{figure*}[ht]
  \centering
   \includegraphics[width= 0.85\textwidth]{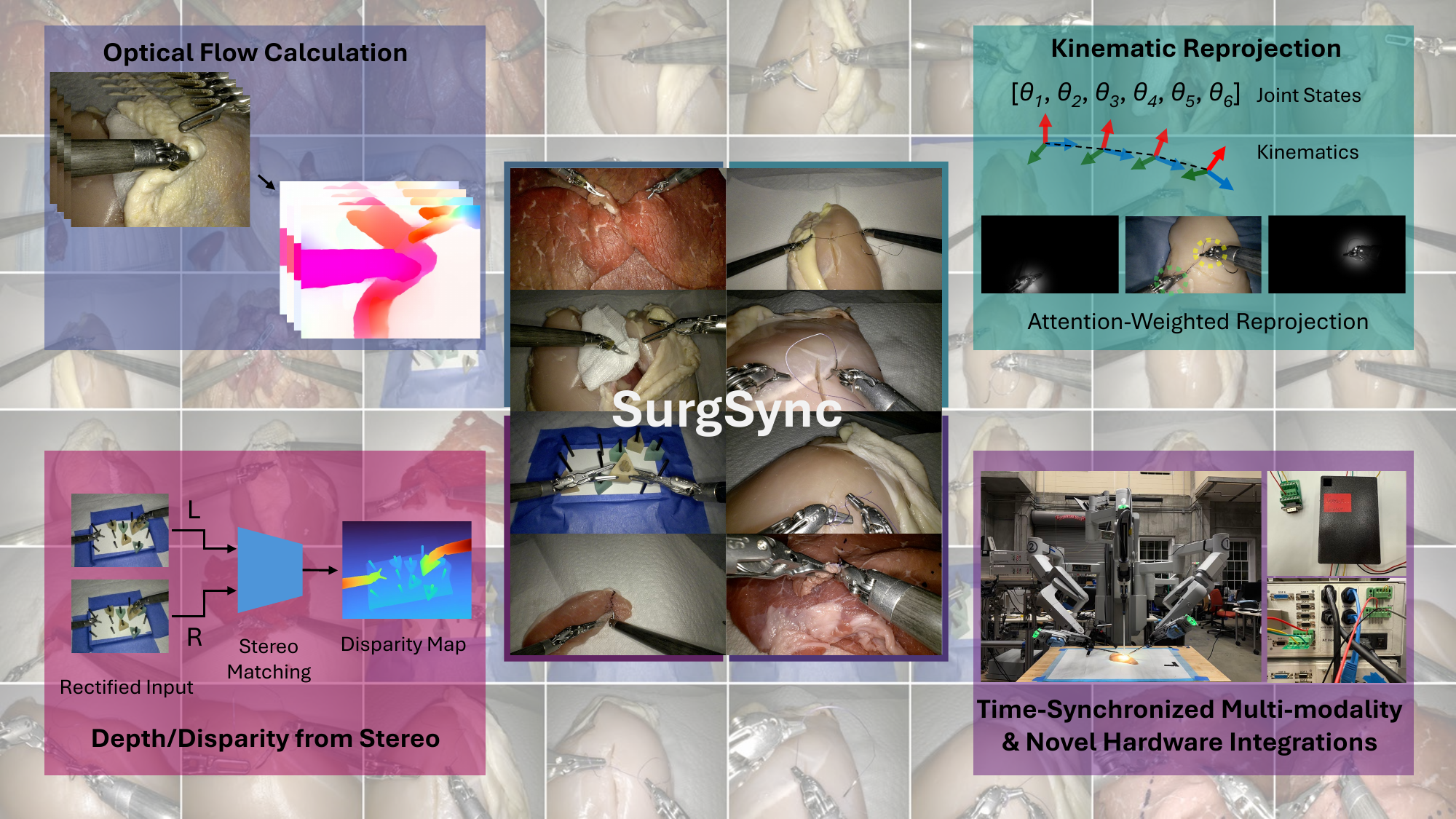}%
  \\
  \includegraphics[width= 0.85\textwidth, trim=15 5 20 0, clip]{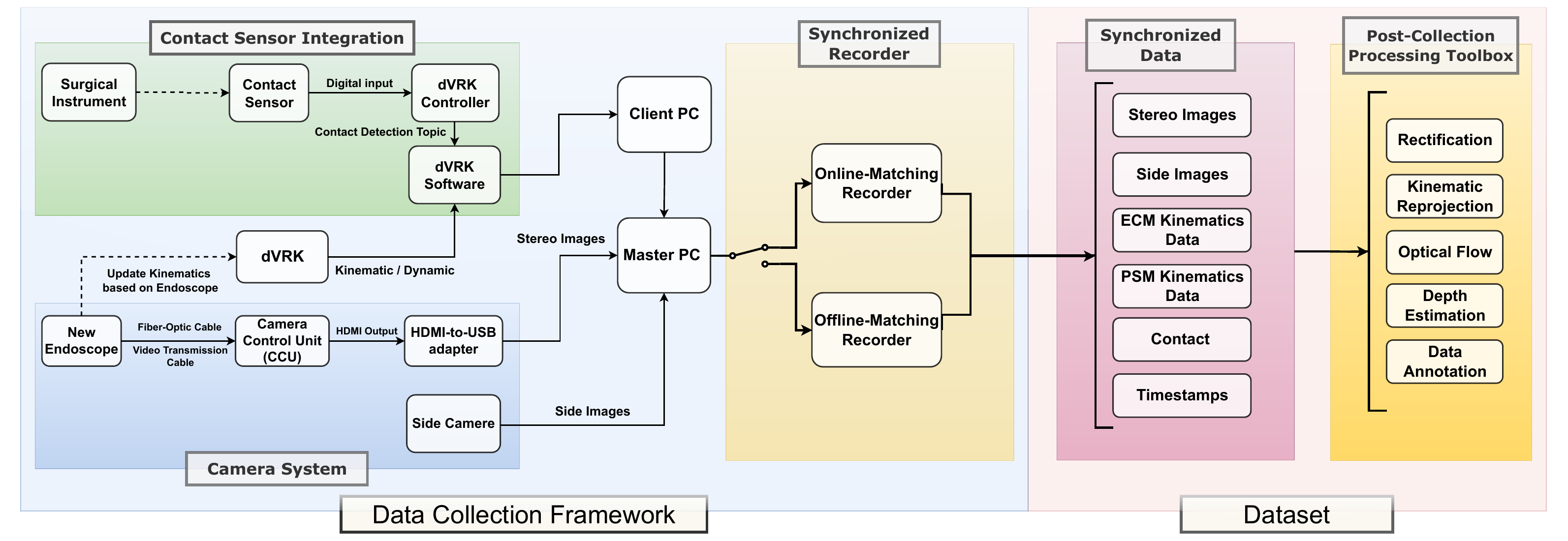}
  \caption{Overview of the proposed SurgSync dataset and data collection framework. We collect 214 recordings while performing multiple canonical training tasks on ex-vivo tissues (primarily) and phantoms. We also implement post-collection processing using our toolbox. Multiple input modalities, including visual, kinematics/dynamics, tool-tissue contact, event/phase description, are provided for further algorithm training.}
  \label{fig:setup_block_plot}
  \vspace{-12pt}  
\end{figure*}

Robot-assisted surgery (RAS) has revolutionized the field of medical science by providing surgeons with enhanced dexterity, advanced visualization, and precision when performing clinical procedures over the past two decades. The success of the da Vinci{\textregistered} Surgical System (dVSS, Intuitive Surgical Inc. Sunnyvale, CA) is the epitome of this revolution \cite{d2021accelerating}. 

In the research domain, high-quality, well-annotated datasets are foundational to progress in artificial intelligence (AI) applications~\cite{o2024open} for RAS \cite{haidegger2022robot}. They enable data-driven perception, modeling and control, spanning instrument tracking \cite{ding2022carts,d2024robust, fernandes2023future, ding2024segstrong, xu2025surgripe, wu2025surgpose}, tissue interaction understanding \cite{yilmaz2020neural, chua2021toward, reyzabal2024dafoes}, skill assessment \cite{varadarajan2009data, zia2018automated, liu2021towards, lam2022machine}, and surgery automation \cite{zhou2024suturing, scheikl2024movement, kim2024surgical, kim2025srt, long2025surgical}.

Despite this momentum, robotics applications, especially for surgical scenarios, face a shortage of high-quality training and validation data due to highly diverse data distribution and the expensive cost of data collection in the physical world. Even though generating synthetic data for training \cite{colleoni2020synthetic, zeng2024realistic, barragan2024realistic, wu2024surgicai, yang2025instrument, wu2025augmenting, moghani2025sufia} can mitigate this problem through sim-to-real transfer, the sim-to-real gap limits the complexity of procedures and overall performance. Therefore, there remains a need for real-world datasets to ensure effective evaluation.

Many available datasets in the surgical robotics domain suffer from three practical limitations: (\romannum{1}) weak or inconsistent time alignment across sensing modalities, which obscures cause-and-effect and degrades sequence models; (\romannum{2}) legacy imaging pipelines that limit visual fidelity and downstream vision performance; and (\romannum{3}) narrow task coverage and post-collection tooling that constrain reproducibility and reuse. These gaps are especially consequential for systems like the da Vinci Research Kit (dVRK, also known as dVRK Classic) \cite{kazanzides2014open} and dVRK-Si \cite{xu2025dvrk}, where fine motor actions, bi-manual coordination, and tissue dynamics evolve on sub-second timescales and must be captured coherently across vision and robot states. 

To address these challenges, we present an open-source data collection framework for surgical robotic systems, such as dVRK Classic and dVRK-Si, with the following contributions:

\begin{itemize}
    \item \textbf{Time-synchronization design pattern:} two synchronized recorders (online/offline-matching) for smooth and continuous teleoperation recording, acknowledging time synchronization as a first-class design constraint;
    \item \textbf{Upgraded imaging stack:} integration of a modern chip-on-tip endoscope (Cornerstone Robotics (CSR) Ltd., Hong Kong, China) with dVRK-Si to enable high-performance imaging stack;
    \item \textbf{Tool-tissue contact ground-truth sensing:} a capacitive contact sensor, interfaced via a digital input of the dVRK controller, for seamless acquisition of tool-tissue contact ground truth
    on ex-vivo tissues;
    \item \textbf{Post-collection processing toolbox:} a configurable and extensible post-collection toolbox for better reusability;
    \item \textbf{User-study dataset:} user studies for data collection, including multiple practical training procedures performed on phantoms and ex-vivo tissues, such as peg transfer, tissue manipulation, suturing and dissection.
\end{itemize}

In addition, we validate the usability of our dataset via implementing a state-of-the-art skill assessment algorithm \cite{liu2021towards} on the suturing-task subset of our data.

Our proposed framework primarily communicates via Robot Operating System (ROS) and can be extended to general surgical robots supported within the ROS ecosystem. Furthermore, the cross-platform validation demonstrates that the framework can be used by multiple groups to create a combined, diverse, large-scale dataset.
An overview of the system and dataset are shown in Fig. \ref{fig:setup_block_plot}.

\section{Related Work}


The surgical robotics domain has been facing a lack of high-quality, well-annotated large-scale public datasets.
Researchers have devoted substantial efforts to generate synthetic data through simulation \cite{colleoni2020synthetic, xu2021surrol, munawar2022open, yu2024orbit, zeng2024realistic, barragan2024realistic, wu2024surgicai, yang2025instrument, wu2025augmenting, moghani2025sufia}, however, a persistent trade-off remains among photorealism, physically faithful tool-tissue interactions and task complexity. Therefore, real-world data remain essential for training and rigorous evaluation.

In 2014, Gao et al. introduced JIGSAWS \cite{gao2014jhu, ahmidi2017dataset}, the first open-source dataset for surgical gesture recognition. It catalyzed AI research in RAS \cite{moglia2021systematic} yet exhibits known limitations \cite{hendricks2024exploring}, including suboptimal image quality and visual fidelity due to the legacy image pipeline.
%
Also in 2014, we introduced the da Vinci Research Kit (dVRK) \cite{kazanzides2014open}, an open-source research platform derived from the dVSS, 
which enabled collection of additional datasets on phantoms for peg transfer \cite{long2025surgical}, tool retraction/palpation \cite{chua2021toward}, pattern cutting \cite{sharon2024augmenting} and other related training tasks \cite{rivas2023surgical}. However, both JIGSAWS and these dVRK datasets were
collected with phantoms, which may not capture the dynamics of ex-vivo or in-vivo tool-tissue interactions.

Beyond these efforts, the EndoVis challenges have released public datasets over several years \cite{fernandes2023future, ding2024segstrong, xu2025surgripe}, providing benchmarks for endoscopic perception and segmentation tasks. Various multi-modal endoscopic datasets \cite{liu2025comprehensive} have also been released for Large Language Model (LLM) investigations. All aforementioned datasets share one or more constraints: (\romannum{1}) suboptimal time alignment across different modalities, (\romannum{2}) narrow task coverage, (\romannum{3}) absence of instrument/robot kinematic data, (\romannum{4}) missing camera parameters for further calibration, and (\romannum{5}) lack of multi-view image information. 

Finally, our previous work \cite{wu2025surgpose} investigated advanced annotation using fluorescence-based approaches, but logged only sparse, discrete events, which limits temporal resolution and can obscure critical dynamics information. 

\section{Methodology}



\subsection{Synchronized Recorder}

We design and implement two practical synchronized recorders in modern C++, which aim to record temporally aligned data streams across multiple sensing modalities. Specifically, they synchronize stereo (or mono) video streams with kinematic data from the patient-side manipulators (PSMs) and endoscopic camera manipulator (ECM) of the dVRK, including both measured and desired (also known as setpoint) states. In addition, the contact sensor signals are seamlessly integrated within the dVRK data stream and recorded with the same time base. We offer two operation modes for the synchronized recorders so that the user can choose between online or offline time-matching approaches. Both recorders operate on smooth and continuous teleoperation, preserving fine-grained temporal context for sequence models, dynamics and interaction analysis. All communications are based on ROS.


\begin{algorithm}[!b]
\small
\caption{Online-Matching Recorder}
\label{alg:online_recorder}
\begin{algorithmic}
\Procedure{OnlineMatchRecorder}{}
    \State \textbf{Initialize Subscribers}
    \State \quad Subscribe to video streams based on user preferences
    \State \quad \textbf{For each} enabled arm (PSMs or ECM):
    \State \quad\quad Subscribe to \texttt{<arm>/measured\_js/cp/cv}
    \State \quad\quad Subscribe to \texttt{<arm>/setpoint\_js/cp}
    \State \quad\quad \textbf{If} arm is PSM: subscribe to jaw streams, contact sensors


    \State \textbf{Data Structures}
    \State \quad Buffers: queues for images, kinematics (per arm)
    \State \quad \texttt{SyncedQueue}: queue of synchronized data packets
    \State \quad Global state: latest setpoint \& jaw values (per arm)


    \State \textbf{Sync Thread}
    \State \quad \textbf{Loop} until cut-off requested from keyboard input:
    \State \quad\quad \textbf{If} \texttt{SyncedQueue} full $\rightarrow$ drop oldest images, continue
    \State \quad\quad \textbf{If} image buffers empty $\rightarrow$ wait
    \State \quad\quad Retrieve image reference timestamp
    \State \quad\quad \textbf{For each} arm:
    \State \quad\quad\quad \texttt{GetClosestFromQueue}(kin\_buffer, ref\_stamp)
    \State \quad\quad Construct \texttt{SyncedPacket} including reference time-
    \State \quad\quad -stamp, images and kinematic information
    \State \quad\quad Enqueue packet to \texttt{SyncedQueue}
    \State \quad\quad Pop used image(s) from buffer
    \State \quad\quad Notify writer threads


    \State \textbf{Writer Threads} (default pool size: $N=4$)
    \State \quad \textbf{Loop:}
    \State \quad\quad Wait on \texttt{SyncedQueue}, pop one packet
    \State \quad\quad Create temporary folder \texttt{<timestamp>}
    \State \quad\quad Save the images as PNGs
    \State \quad\quad Save the arm kinematics to JSON files
    \State \quad\quad (Optional) Save data to HDF5 and convert later


    \State \textbf{Shutdown and Cleanup (Press `q' on the keyboard)}
    \State \quad Obtain the cut-off timestamp $t_{end}$
    \State \quad Release buffers until the reference timestamp passes $t_{end}$
    \State \quad Remove incomplete folders
    \State \quad \texttt{ReformatDataStorage()} into final dataset layout
\EndProcedure
\end{algorithmic}
\end{algorithm}


\begin{algorithm}[!b]
\small
\caption{Offline-Matching Recorder}
\label{alg:offline_recorder}
\begin{algorithmic}
\Procedure{DetachedRecorder}{}
    \State \textbf{Initialize Directory}
    \State \quad Create \texttt{<run\_id>} with subdirs: \texttt{kin/}, \texttt{meta/}
    \State \textbf{Initialize Subscribers}
    \State \quad Subscribe to video streams based on user preferences
    \State \quad \textbf{For each} enabled arm (PSMs or ECM):
    \State \qquad Subscribe to \texttt{<arm>/measured\_js/cp/cv}
    \State \qquad Subscribe to \texttt{<arm>/setpoint\_js/cp}
    \State \qquad \textbf{If} PSM: subscribe to jaw streams, contact sensors
    \State \textbf{Video Writer} (fixed-rate)
    \State \quad Create \texttt{VideoStreamRecorder(s)} for left/right[/side]
    \State \quad Record frames at pre-defined FPS
    \State \quad \textbf{If} ahead of schedule:
    \State \qquad Wait until next frame's expected timestamp
    \State \quad Image callbacks push frames to corresponding recorder(s)
    \State \textbf{Kinematic Writer} (binary)
    \State \quad \textbf{For each} arm/topic callback:
    \State \qquad Append to \texttt{FastBinWriter} file in \texttt{kin/}
    \State \textbf{Runtime Control}
    \State \quad Start AsyncSpinner for callbacks
    \State \quad Launch key listener; press `q' $\rightarrow$ stop recording
    \State \textbf{Shutdown}
    \State \quad Stop recorders; flush and close all writers
    \State \quad Write \texttt{start\_times.json} and \texttt{end\_times.json}
\EndProcedure
\Procedure{Detached Post-Collection Matching}{}
    \State Decouple the videos into independent image frames
    \State Convert binary kinematic files to readable JSON files
\EndProcedure
\end{algorithmic}
\end{algorithm}


\subsubsection{Online-Matching Recorder} 


The design enforces strict time synchronization using multi-threading: only samples that fall within a user-defined time tolerance are admitted. This yields slightly uneven inter-sample intervals (irregular $\Delta t$), but retains the natural continuity of smooth teleoperation segments and avoids label/feature drift. In our study, a time tolerance of 10\,ms is selected to ensure both tight time alignment and consecutive recorder output. The choice of the time tolerance depends on task requirements and hardware I/O performance. This design can be used for real-time scenarios. 




\subsubsection{Offline-Matching Recorder}


Our offline-matching approach decouples recording from time alignment to maximize the recording system efficiency. Therefore, this recorder produces synchronized datasets in two stages: (\romannum{1}) a lightweight recorder logs camera streams to videos and raw kinematic streams to binary files with minimal processing; (\romannum{2}) an offline post-processing pipeline reconstructs a fixed-rate frame sequence and, for each frame, gathers the $k$ closest samples using nearest-timestamp lookup for subsequent interpolation (we select $k=1$ for simplicity). Compared to the online-matching recorder (which pairs visual and kinematic data in real time), this two-stage design avoids tolerance-based dropping of data during capture, yielding a higher throughput and uniform intervals between synchronized data packets at the cost of requiring more storage and substantial time for post-collection time-matching and interpolation.

This offline-matching recorder is intended for data collection where real-time data acquisition is non-essential, for example, building large training datasets for robot policy learning. By recording every camera frame at the target FPS and interpolating kinematics offline using a configurable rule, the pipeline eliminates tolerance-based drops at capture and produces a uniformly sampled training set. It is the more suitable choice for offline learning workflows, where deterministic post-processing is preferable to on-the-fly data matching.

\subsection{Contact Sensor Integration}

\begin{figure}[ht]
  \centering
  \includegraphics[width=0.85\linewidth]{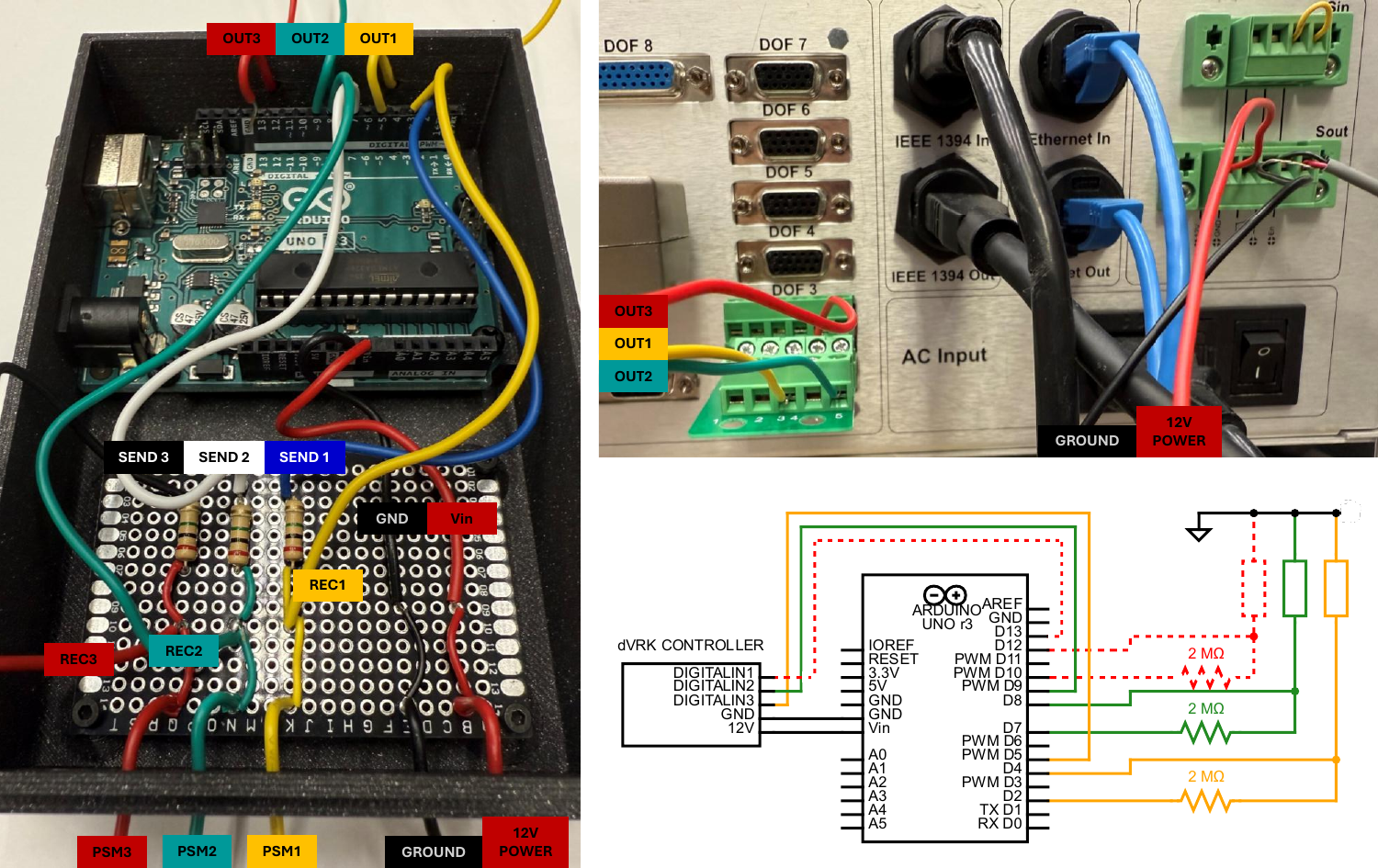}
  \caption{Hardware setup and schematics. The Arduino and the proto-board are shown on the left, the schematics of the electronics on the bottom right, and the connection to the dVRK controller on the top right. }
  
  \label{fig:schematics}
\end{figure}

Our framework supports the integration of custom sensors. For example, we 
incorporated a contact sensor implemented with an Arduino UNO Rev3 and the Capacitive Sensing Library \cite{capacitivesensinglibrary}. 
The assembled hardware prototype and schematics are shown in Fig. \ref{fig:schematics}. 


\begin{figure}[ht]
  \centering
  \includegraphics[width=0.9\linewidth]{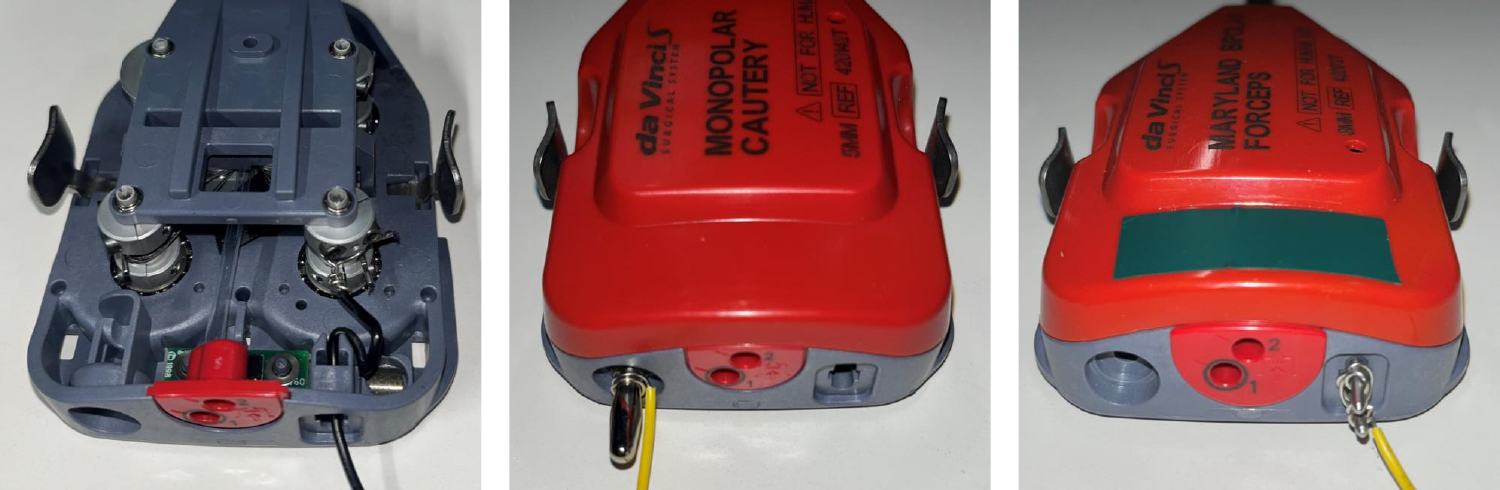}
  \caption{Instrument wire connection. Non-polar instrument (left), monopolar (center), and bipolar (right). For non-polar instrument, the tool housing is opened and the wire insulation layer is removed for better visualization. }
  \label{fig:instrument_connection}
\end{figure}

For monopolar and bipolar instruments, the connection to the sensor is achieved by wrapping a wire around their connectors; for non-electrosurgical (non-polar) instruments, we opened the tool housing, and attached a wire to the rotation coupling of one joint with proper strain relief and wire insulation (Fig. \ref{fig:instrument_connection}). 

The library computes the capacitance at the receiving pin in arbitrary units; the returned values depend on the chosen resistor and on the material and size of the contacting object. The sensor exhibits the best performance with human tissue, while ex-vivo animal tissue such as chicken breast, thin-sliced beef or pork also provide similarly reliable results. A contact threshold is defined to binarize the signal into contact and non-contact states, which are then transmitted through the Arduino output pin to the digital input of the dVRK controller (Fig. \ref{fig:schematics}, top right). Each digital input is registered in the dVRK software framework by specifying bit ID, FPGA board, and related parameters in a configuration file.

In this study, we used 2\,M$\Omega$ resistors and a threshold of 205 for the best performance on the ex-vivo tissues used in our experiments. To evaluate the contact sensor performance, we randomly select one instance each from tissue manipulation, suturing and dissection, as shown in section \ref{sec:user_study}, and assess detection accuracy. The resulting accuracies are 99.1\% for tissue manipulation, 74.3\% for dissection and 45.2\% for suturing. The misclassifications are primarily attributed to sensor noise, humidity-induced changes in capacitance and intermittent short-circuit events when instruments contacted the same conductive object (e.g., a suturing needle or a small tissue fragment). Those misclassifications are eventually addressed by manual re-annotation using the GUI in section \ref{sec:data_gui}.

\subsection{Modern Endoscope Integration}


We integrate a contemporary chip-on-tip endoscope with the dVRK-Si system to improve visual fidelity and enable frame-accurate alignment with robot kinematics and contact signals. The full imaging pipeline is shown in Fig.~\ref{fig:setup_block_plot} (bottom), comprising hardware mounting, signal capture, and ROS integration. Our integration yields higher-quality images compared to the default dVRK-Si scope and enables coherent multi-modal dataset construction.

\begin{figure}[!b]
  \centering
  \includegraphics[width=0.9\linewidth]{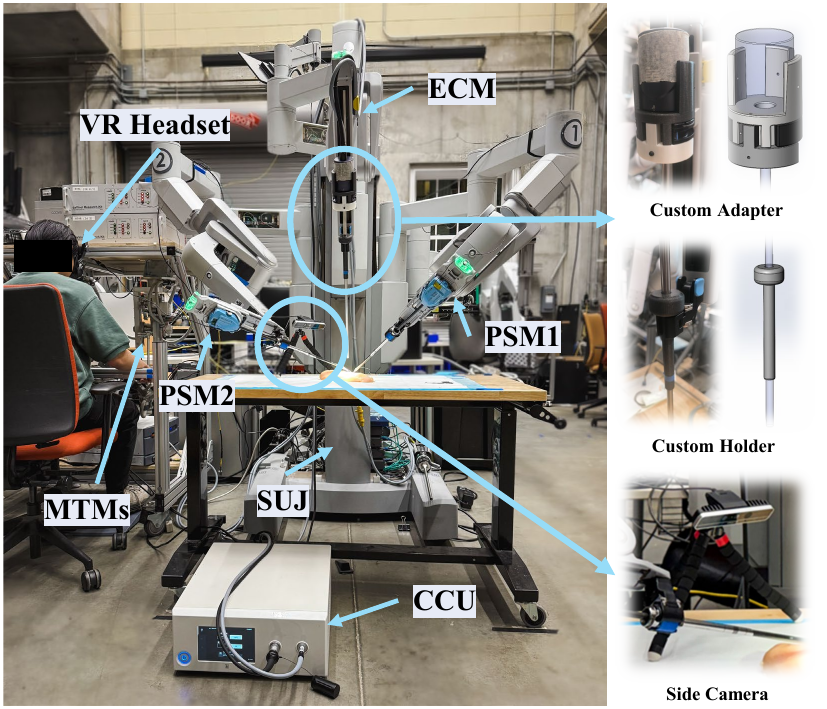}
  \caption{Overall experimental setup. The endoscope is mounted on the dVRK-Si ECM using a custom holder and adapter. }
  \label{fig:setup}
\end{figure}

\subsubsection{Hardware Integration}
The endoscope connects to a clinical-grade Camera Control Unit (CCU) via video and fiber-optic illumination cables. The CCU outputs two 1080p HDMI video streams, which are routed to the host PC using HDMI-to-USB frame capture devices. A compact 3D-printed holder mounts the endoscope onto the dVRK-Si Endoscopic Camera Manipulator (ECM) with coaxial alignment, quick attachment, and strain relief. CAD models and overall system setup are shown in Fig.~\ref{fig:setup}.

\subsubsection{Software Integration}
On the host PC, frames are acquired through the \texttt{v4l2src} backend using \texttt{gscam} \& \texttt{gstreamer} and published to ROS image topics. Each frame is timestamped on arrival using the host monotonic clock. This shared time base enables precise temporal alignment during post-processing. The same pipeline supports stereo or side-view cameras through multiple synchronized ROS nodes.
 
\subsubsection{ Image Quality Comparison}

\begin{figure}[ht]
  \centering
  \includegraphics[width=0.45\linewidth]{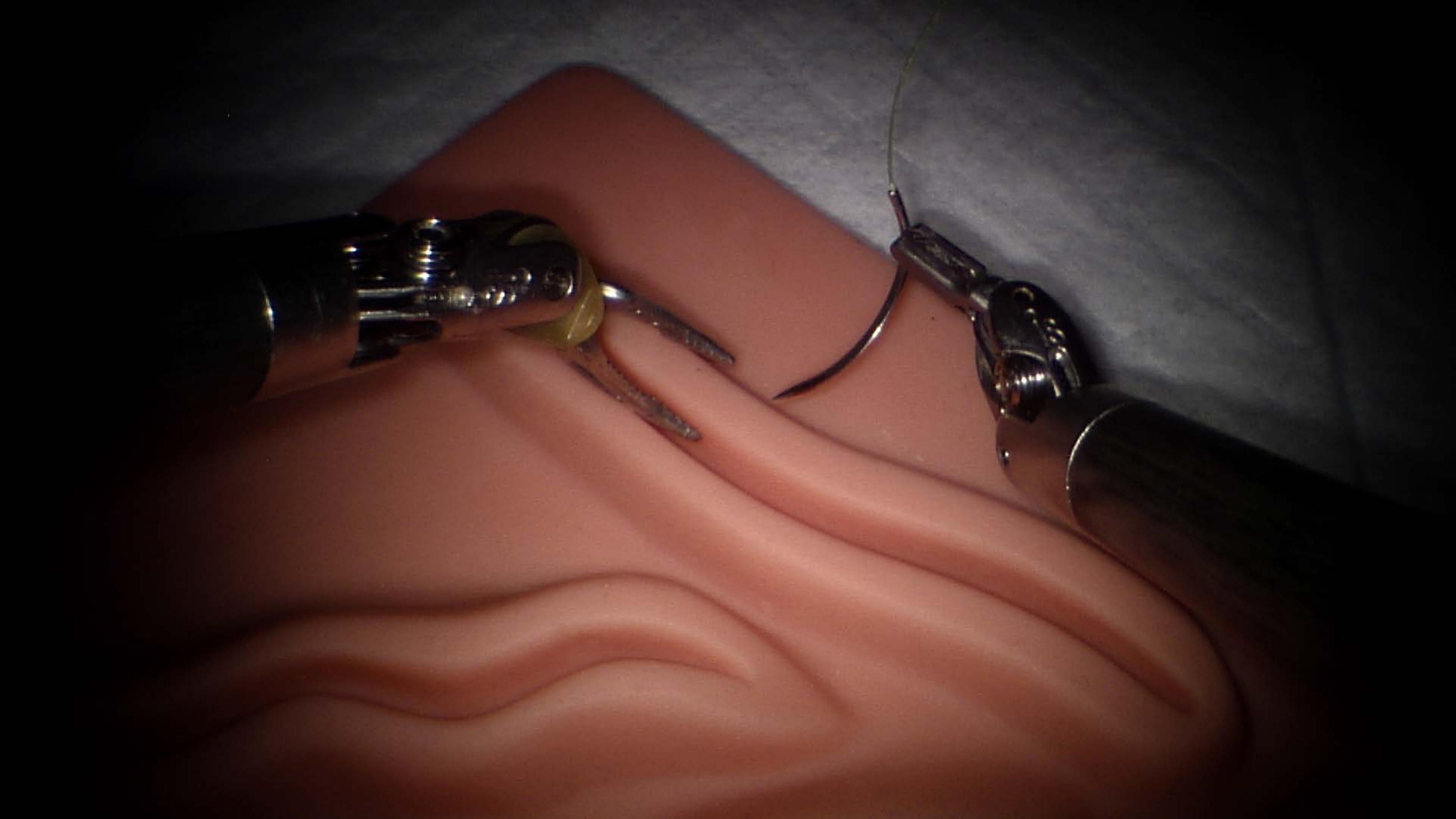}
  \includegraphics[width=0.45\linewidth]{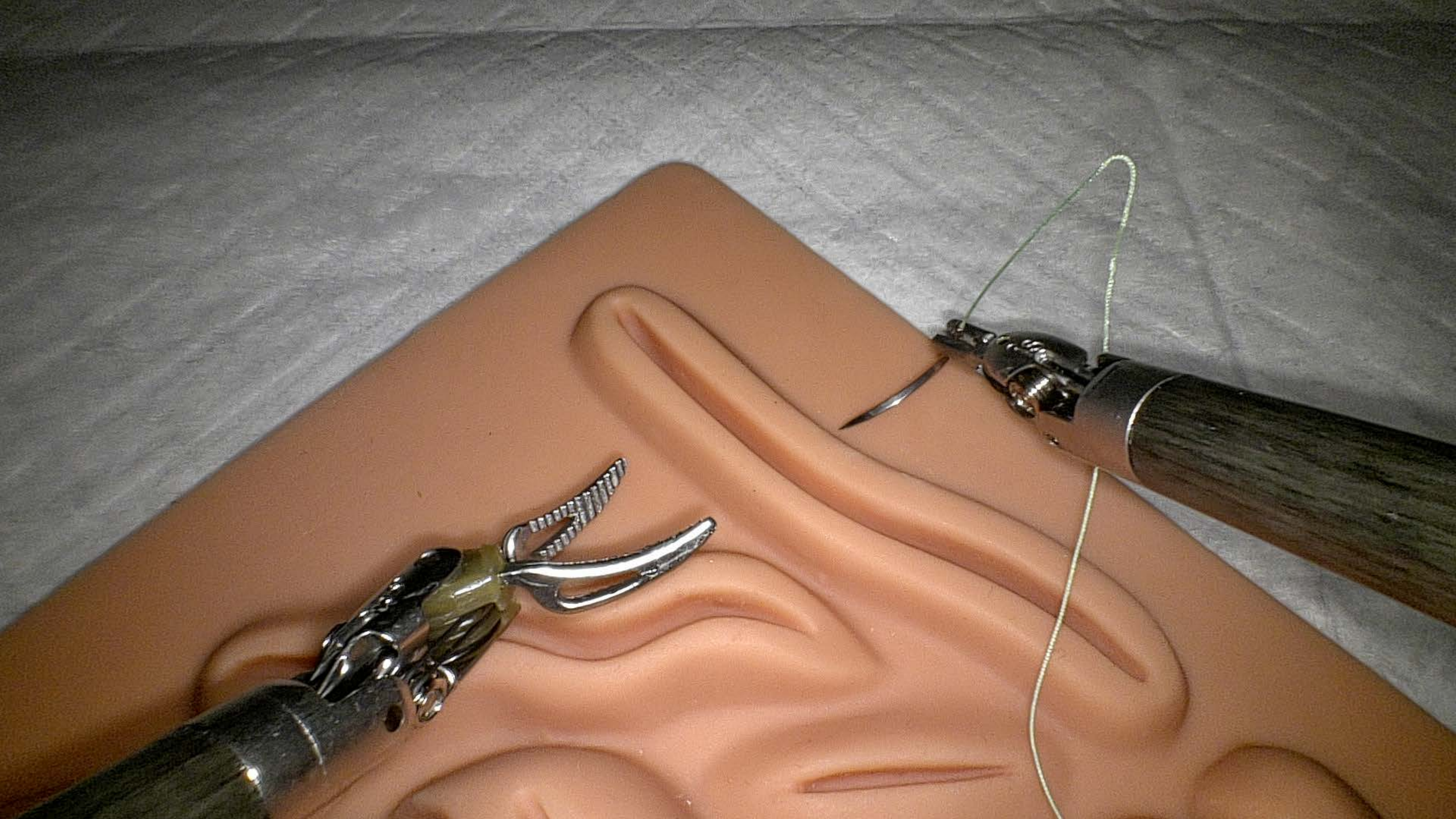}
    \caption{Representative frames from the dVRK-Si endoscope (left) and our modern CSR endoscope (right). \emph{Brightness settings:} dVRK-Si CCU at 100\%, ours at 30\%. Both frames are captured with similar camera-phantom distances.}
  \label{fig:img-quality-comparison}
\end{figure}

Compared to the legacy dVRK-Si endoscope, our integrated imaging system yields significantly sharper visual frames. Quantitatively, the average Laplacian variance~\cite{pertuz2013analysis} is over 30$\times$ higher in our system as shown in Table~\ref{tab:laplacian-results}, indicating substantially improved details and edge clarity. 
These improvements benefit downstream perception tasks such as segmentation, optical flow calculation, and depth estimation. Representative frames from each system are shown in Fig.~\ref{fig:img-quality-comparison}. 

\begin{table}[tbh]
  \caption{Laplacian variance on suturing sequences.}
  \label{tab:laplacian-results}
  \centering
  \setlength{\tabcolsep}{6pt}
  \sisetup{table-number-alignment = center, round-mode=places, round-precision=2}
  \begin{tabular}{
      c
      S[table-format=3.2] 
      S[table-format=2.2]  
  }
    \hline
    \textbf{System} & {\textbf{Mean} } & {\textbf{STD} } \\
    \hline
    CSR Endoscope (Ours) & 529.48 & 23.77 \\
    dVRK-Si Endoscope   &  16.93 &  2.47 \\
    \hline
  \end{tabular}
\end{table}

\subsection{Post-Collection Processing Toolbox}

Beyond data acquisition, we provide an extensible and highly configurable post-collection processing toolbox that standardizes calibration and data processing. It supports stereo rectification, image resizing, projection of robot kinematics into the endoscopic image frames, and generation of derived modalities, including disparity/depth and optical flow. In addition, it also produces comprehensive annotations for contact detection and event/phase labels. 

\subsubsection{Kinematic Reprojection}

We propose a practical approach using a Gaussian heatmap to project tool tip 3D position kinematic information~\cite{lee2020camera} to 2D gray-scale endoscopic images.
This relies on a hand-eye calibration \cite{dvrk_hand_eye} to correct for the well-known inaccuracy of the dVRK kinematics \cite{cui2023caveats}.

\begin{figure}[th]
    \centering
    \includegraphics[width=0.95\linewidth]{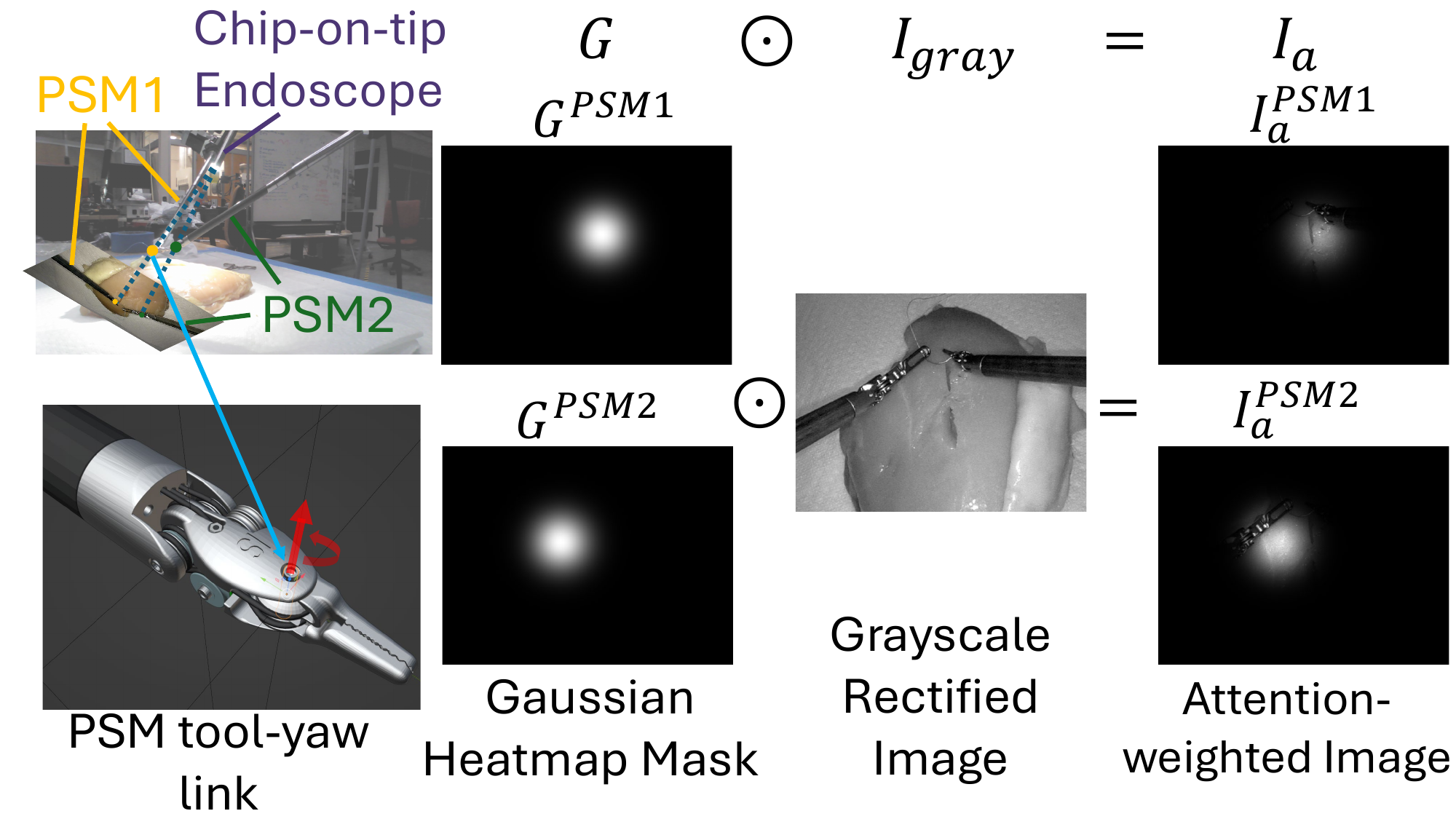}
    \caption{Kinematic Reprojection Pipeline. This approach projects the Cartesian positions of the PSM tool-yaw link to attention-weighted images.}
    \label{fig:kinematic_project}
\end{figure}

Given the hand-eye calibration and stereo camera parameters, the 3D point $p = (x_p,y_p,z_p)^T$ on the dVRK PSM tool-yaw link (shown in Fig.~\ref{fig:kinematic_project}) is projected to the image plane $(u_p, v_p)$ using a pinhole camera model. We can then generate a Gaussian heatmap $G$ centered at ($u_p$, $v_p$):
\begin{equation}
        G(p_x, p_y) = e^{-\left ( \frac{(p_x - u_p)^2}{\sigma_x^2} +\frac{(p_y - v_p)^2}{\sigma_y^2} \right )}
    \label{eq:heatmap}
\end{equation}
where $(p_x,p_y)$ indexes image pixels and $\sigma_x$, $\sigma_y$ control the spread of the heatmap along the image axes. 

Eventually, we compute an element-wise product between the rectified grayscale stereo images ($I_{gray}$) and the heatmap mask $G$ to obtain attention-weighted images ($I_a = G \odot I_{gray}$) that emphasize the region of interest for further interaction analysis. The overall pipeline is shown in Fig.~\ref{fig:kinematic_project}.

\subsubsection{Data Annotation}
\label{sec:data_gui}

\begin{figure}[ht]
    \centering
    \includegraphics[width=0.95\linewidth]{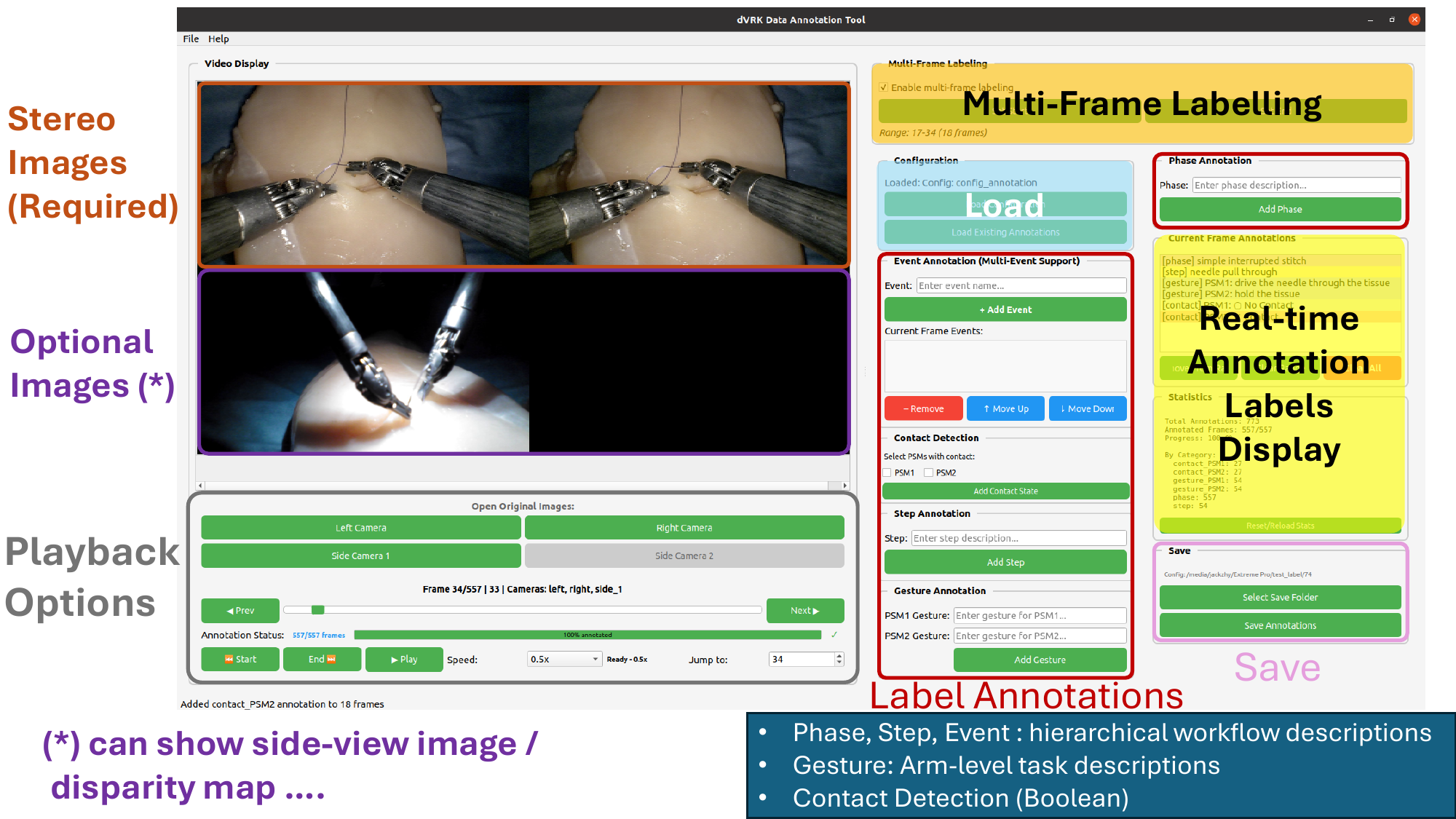}
    \caption{Data Annotation GUI}
    \label{fig:data_gui}
\end{figure}

We developed a custom data annotator with Graphical User Interface (GUI) using PyQt as shown in Fig.~\ref{fig:data_gui}, enabling users to label contact detection and event/phase description during playback of consecutive frames.

\subsubsection{Depth Estimation}

As shown in Fig.~\ref{fig:setup_block_plot}, we perform disparity estimation using the FoundationStereo model~\cite{wen2025foundationstereo} and obtain disparity images. Furthermore, the disparity can be converted to depth via $depth = \frac{f \times b }{disparity}$, where $f$ and $b$ represent the focal length and baseline distance of the stereo endoscope that can be directly obtained from the stereo camera parameters.

\subsubsection{Optical Flow}

As shown in Fig.~\ref{fig:setup_block_plot}, we also employ the RAFT model~\cite{teed2020raft} to compute dense optical flow between consecutive frames. We then apply a custom magnitude-aware filter to suppress low-magnitude noise for better image output.

\section{Experiment Setup}

\subsection{System Configuration}

Fig.~\ref{fig:setup} shows the primary system configuration at Johns Hopkins University (JHU). To assess cross-platform generality, we deployed a second configuration on the dVRK Classic at the University of British Columbia (UBC). Both systems share the same camera architecture (stereo endoscope + side-view camera), although the UBC system uses the legacy dVRK Classic endoscope rather than the modern unit and omits the contact sensor. In the UBC setup, cameras stream to a client workstation, while the I/O-intensive dVRK software framework runs on the master workstation. The hand-eye calibration was not performed at UBC. Both setups use an Intel\textregistered{}\,RealSense$^{TM}$ RGBD camera as the side-view camera.

Both recorder systems run on workstation-class CPUs: Intel\textregistered{}\,Xeon(R) W-2245 (JHU) and Intel\textregistered{}\,Core$^{TM}$ i9-11900 (UBC). All cameras produce raw image streams with a resolution of 1080p and 30\,Hz (side-view) or 60\,Hz (stereo endoscope) refresh rate. The dVRK framework runs at 1\,kHz. At JHU, the offline-matching recorder sustains up to 10\,Hz and omitting the side-view camera can increase the rate to 15\,Hz.

\subsection{User Study}
\label{sec:user_study}

We conduct a user study to collect data for multiple canonical training tasks on phantoms or ex-vivo tissues, which include but are not limited to:

\begin{itemize}
    \item Peg transfer (phantoms and gauze in chicken breast)
    \item Single interrupted suturing practice (chicken breast)
    \item Tissue manipulation (chicken hearts/breast, beef and pork)
    \item Dissection following a trace (beef and pork)
\end{itemize}

13 human subjects (3 female, 10 male) participated in our user study. 4 subjects have limited knowledge of dVRK operation and are considered novice users (N); 5 are familiar with dVRK operation and are considered experienced users (E); 4 are surgeons and are considered professional users (P).



\section{Results}


\subsection{Cross-Platform Validation}

The open-source data collection framework successfully executed on two different setups at two different institutions (JHU and UBC). The UBC setup achieved similar performance using a larger online-matching time tolerance of 100\,ms due to the heavy I/O load from the dVRK software framework. 

\subsection{Dataset Distribution}

We collected 214 instances over multiple training tasks in our user study. The total frame count per instance varies, depending on the task and the user's skill level. Of the 214 validated instances, 9 were collected on the UBC setup and only used the online-matching recorder without the hand-eye calibration. 102 instances were collected using the offline-matching recorder. 96 offline-matching instances were collected within the Intuitive abdominal dome. A brief dataset distribution with respect to training tasks and human subjects' skill level is shown in Table~\ref{tab:dataset_detail}.

\begin{table}[tbh]
    \centering
    \caption{Dataset Distribution, User Groups defined in section \Romannum{4}-B}
    \begin{tabular}{| c | c | c | c | c | c |}
         \hline
         \multirow{2}{*}{\makecell{\textbf{Training} \\ \textbf{Task}}} & \multirow{2}{*}{\makecell{\textbf{User} \\ \textbf{Group}}} & \multicolumn{2}{c|}{\textbf{Number of Instances}} & \multirow{2}{*}{\textbf{Total}} \\
         \cline{3-4}
          &  & \textbf{Online} & \textbf{Offline} & \\
         \hline
         \multirow{3}{*}{\makecell{Suturing \\ and \\ Knot Tying}} & N & 13 & 2 & \multirow{3}{*}{\makecell{104}} \\ 
         \cline{2-4}
          & E & 36 & 12 & \\
          \cline{2-4}
          & P & 2 & 39 & \\
         \hline
         \multirow{2}{*}{\makecell{Peg Transfer}} & N & 7 & \multirow{2}{*}{-} & \multirow{2}{*}{18} \\ 
         \cline{2-3}
         & E & 11 &  & \\
         \hline
         \multirow{2}{*}{\makecell{Tissue Manipulation}} & N & 9 & \multirow{2}{*}{-} & \multirow{2}{*}{21} \\ 
         \cline{2-3}
         & E & 12 & & \\
         \hline
         \multirow{3}{*}{\makecell{Dissection}} & N & 6 & - & \multirow{3}{*}{71} \\ 
         \cline{2-4}
          & E & 15 & 9 & \\
          \cline{2-4}
          & P & 1 & 40 & \\
         \hline
    \end{tabular}
    \label{tab:dataset_detail}
\end{table}

\subsection{Synchronized Recorder Evaluation}

To evaluate time-synchronization performance, we randomly selected 4 instances for each recorder from the JHU dataset. The online-matching recordings contain 1523 frames in total and the offline-matching recordings contain 4876 frames. For each frame, excluding the contact sensor modalities which are latched topics, the recorded data contains 20 time-synchronized ROS topics. For the offline-matching recorder, we use all five closest samples for the calculation.

\begin{figure}[!b]
\vspace*{-\baselineskip}

\begin{minipage}{\columnwidth}

\begin{table}[H]
  \centering
  \caption{Synchronized Recorder Comparison and Time Analysis}
  \footnotesize
  \begin{tabular}{|c|c|c|}
    \hline
    \textbf{Attribute} &
    \textbf{\makecell{Online- \\ matching}} &
    \textbf{\makecell{Offline- \\ matching}} \\
    \hline
    \makecell{Time latency \\ mean $\pm$ std (ms)} & 6.36 $\pm$ 4.72 & 1.35 $\pm$ 0.81 \\
    \hline
    \makecell{Time latency \\ median (ms)}       &  5.58          & 1.33          \\
    \hline
    \makecell{Recording \\ frequency (Hz)}  & 4.04 $\pm$ 1.69       & \makecell{10}     \\
    \hline
    Ready to use  & Yes           & \makecell{No \\ (post-collection time matching \\ and interpolation required)}            \\
    \hline
  \end{tabular}
  \label{tab:recorder_compare}
\end{table}

\end{minipage}
\begin{minipage}{\columnwidth}

\begin{figure}[H]
  \centering
  \includegraphics[width=0.95\linewidth, trim=10 0 20 10, clip]{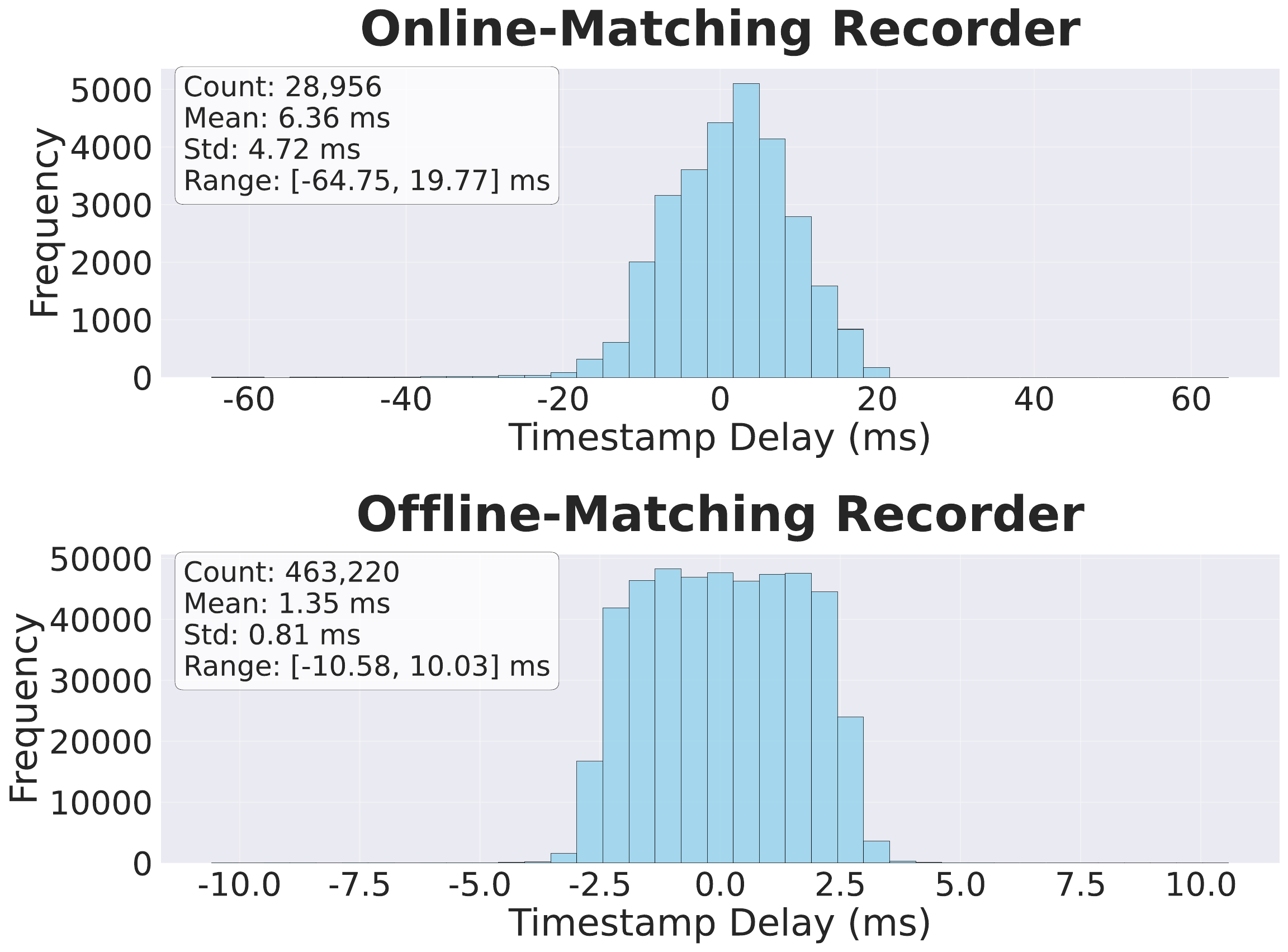}
  \caption{Overall raw time latency distribution of all ROS topics}
  \label{fig:syn_analysis1}
\end{figure}

\end{minipage}
\end{figure}

For both recorders, we select the ROS image topic timestamp of the left stereo endoscope camera as the reference timestamp to perform time latency analysis for synchronization quality evaluation across different modalities. Table~\ref{tab:recorder_compare} summarizes descriptive statistics
and
Fig.~\ref{fig:syn_analysis1} shows the raw time latency distribution. The few outliers could be due to: (1) the time latency for the control loop of the dVRK software, (2) extra time to numerically solve inverse kinematics near singularities, and (3) suboptimal CPU performance.

The stereo camera parameters are included in the dataset. Notably, our post-collection processing tool can also properly handle data subsets involving endoscope movements if the hand-eye calibration is performed.


\subsection{Dataset Validation in Skill Assessment}

We evaluate our synchronized data on the task of skill assessment to demonstrate the feasibility of the collected data and the post-collection processing toolbox. We follow a data-driven regression approach using the unified multi-path framework for automatic surgical skill assessment \cite{liu2021towards}.

After interpolating the second-stage outputs from the offline-matching recorder, the resulting dataset shares identical data structure with the online-matching one. We selected 43 instances from 8 users with varying experience levels. Each instance is split into suturing or knot-tying segments based on manual annotations.


An experienced user with knowledge of animal surgery graded all instances using the global rating score (GRS) \cite{liu2021towards} in the range of 6 to 30. The GRS is derived from an objective rubric~\cite{martin1997objective} with six categories: (1) respect for tissue, (2) suture/needle handling, (3) time and motion, (4) flow of operation, (5) quality of final product, and (6) overall performance. Each category is scored on a scale of 1–5.

The model takes three kinds of synchronized input modalities: (i) kinematic features (14D), including Cartesian positions (6D), measured velocities (6D), and gripper openings (2D); (ii) visual features (2048D) extracted from a ResNet-101 encoder applied to RGB images; and (iii) one-hot encoded gesture labels (14D) based on 14 common surgical gestures, following the JIGSAWS gesture taxonomy \cite{gao2014jhu}. All modalities are aligned per-frame using our synchronized recorder to ensure coherent temporal context across streams.

To ensure diverse coverage of skill levels, we construct four stratified folds for cross-validation~\cite{tang2020uncertainty}. Each fold partitions data into training and test sets with a balance of GRS distribution and user identity. 

Our model follows the multi-path design of Liu et al. \cite{liu2021towards}, with temporal encoders for kinematic (T), visual (V), and gesture (E) streams. Each head outputs the five GRS scores and is trained with equal-weight mean squared error on normalized labels. A temporal contrastive loss is added for regularization. We report Spearman’s rank correlation coefficient (SROCC) averaged over four folds (Table \ref{tab:skill-results}).



\begin{table}[ht]
\caption{Skill assessment performance on our synchronized dataset. SROCC is reported for each cross-validation fold.}
\label{tab:skill-results}
\centering
\setlength{\tabcolsep}{6pt}
\renewcommand{\arraystretch}{1.15}
\begin{tabular}{lcccc|c}
\hline
\textbf{Task} & \textbf{Fold 1} & \textbf{Fold 2} & \textbf{Fold 3} & \textbf{Fold 4} & \textbf{Mean~$\pm$~Std} \\
\hline
Suturing    & 0.658 & 0.828 & 0.844 & 0.881 & \textbf{0.803~$\pm$~0.086} \\
Knot Tying  & 0.870 & 0.875 & 0.618 & 0.699 & \textbf{0.765~$\pm$~0.111} \\
\hline
\end{tabular}
\end{table}

\section{Discussion and Conclusion}

In this manuscript, we propose SurgSync, a multi-modal data collection framework with (1) dual-mode (online/offline) time-synchronized recorder, (2) advanced hardware integration of a modern stereo endoscope and a novel contact sensor on the dVRK-Si, and (3) a post-collection toolbox for rectification, depth estimation and a kinematic reprojection approach using a Gaussian heatmap. We conduct a user study to acquire datasets and demonstrate feasibility by training and evaluating a skill-assessment model on the collected data. The performance of the synchronized recorders is highly dependent on the workstation hardware, yet it can still obtain reliable data even with suboptimal hardware. All open-source software and data are available at \url{https://surgsync.github.io/}

In future work, we will perform additional data collection using the offline-matching approach to build a large-scale training dataset. Moreover, we will include additional data with more types of instruments and training tasks and add experiments with in-vivo or cadaver environments if possible.





\section*{ACKNOWLEDGMENT}

Thanks to Dale Bergman, Alessandro Gozzi and the Intuitive Foundation for all the hardware support of the dVRK-Si. Thanks to Cornerstone Robotics Ltd. for providing the modern chip-on-tip endoscope.




\bibliographystyle{IEEEtran}
\bibliography{references.bib}


\end{document}